\pdfoutput=1
\documentclass[11pt]{article}

\usepackage[]{acl}

\usepackage{times}
\usepackage{latexsym}
\usepackage{graphicx}
\usepackage{booktabs}
\usepackage{adjustbox}
\usepackage[T1]{fontenc}
\usepackage[utf8]{inputenc}

\usepackage{microtype}

\usepackage{inconsolata}

\usepackage{soul}
\usepackage{xcolor}

\usepackage{multirow}
\def\ours{\texttt{TTG}}
\def\oursfull{To the Globe }

\title{\oursfull (\ours): Towards Language-Driven Guaranteed Travel Planning}

\newcommand\tinytriangle{\vcenter{\hbox{\scalebox{0.5}{$\triangle$}}}}

\author{Da Ju$^{\tinytriangle, *}$ Song Jiang$^{*}$  Andrew Cohen$^{*}$ Aaron Foss$^{+}$ Sasha Mitts$^{+}$ Arman Zharmagambetov\\\bf Brandon Amos\ \ \  Xian Li\ \ \   Justine Kao\ \ \   Maryam Fazel-Zarandi\ \ \  Yuandong Tian$^{\tinytriangle, *}$ \\
$\tinytriangle$ Project lead\ \ \ \ \ \ $*$ Core contributor\ \ \ \ \ \ $+$ Equal contribution\\
Meta AI (FAIR)
}

\begin{document}
\maketitle

\begin{abstract}
Travel planning is a challenging and time-consuming task that aims to find an itinerary which satisfies multiple, interdependent constraints regarding flights, accommodations, attractions, and other travel arrangements. In this paper, we propose \emph{\oursfull}(\ours{}), a real-time demo system that takes natural language requests from users, translates it to symbolic form via a fine-tuned Large Language Model, and produces optimal travel itineraries with Mixed Integer Linear Programming solvers. The overall system takes $\sim 5$ seconds to reply to the user request with guaranteed itineraries. To train \ours{}, we develop a synthetic data pipeline that generates user requests, flight and hotel information in symbolic form without human annotations, based on the statistics of real-world datasets, and fine-tune an LLM to translate NL user requests to their symbolic form, which is sent to the symbolic solver to compute optimal itineraries. Our NL-symbolic translation achieves $\sim 91\%$ exact match in a backtranslation metric (i.e., whether the estimated symbolic form of generated natural language matches the groundtruth), and its returned itineraries have a ratio of $0.979$ compared to the optimal cost of the ground truth user request. When evaluated by users, \ours{} achieves consistently high Net Promoter Scores (NPS) of $35$-$40\%$ on generated itinerary. 

\iffalse
\ours{} contains three main components: 1) a \emph{Travel Generator} which generates user requests in symbolic form, 2) a Translator that converts natural language requests to its symbolic form, and 3) a \emph{Travel Solver} which solves the underlying mixed integer linear programming (MILP) optimization problem to yield optimal solutions (i.e., itinerary), and return them to the users. We train the Instruction Creator and Translator using synthetic data. The generation process leverages an existing travel dataset plus simple prompting, without human annotations. Compared to existing pure LLM-based systems, \ours{} provides an up-to-date, precise and executable itinerary with optimality guarantees, in seconds. 
\fi
\end{abstract}

\section{Introduction}
Travel planning is a routine activity that typically requires a significant amount of human time and effort to find an optimal itinerary satisfying many implicit and explicit constraints which interact and change over time. Ideally, a human would only need to provide natural language instructions (e.g., \emph{``I want to go to Hawaii for three days with a budget of \$1,000.''}) and an AI agent provides solutions which are optimal with respect to certain objectives (e.g., total expense) and feasible (i.e., satisfy all constraints). Moreover, the quality of the agent's decision should be reliable enough such that humans can fully delegate the task, or approve with a glimpse of check.  

Designing such an AI system remains non-trivial. First, it involves \emph{complex planning} with potentially vague natural language instructions, sophisticated objectives and constraints (e.g., hotels, flights, restaurants, attractions, budgets) and requiring multiple back-and-forth reasoning steps without a clear and predefined decision path. Despite impressive performance achieved by Large Language Models (LLMs), they are still weak at complex reasoning and planning~\cite{valmeekam2022large,xie2024travelplanner,huang2023large}, and may hallucinate~\cite{huang2023survey} or be inconsistent~\cite{huang2023large}, in particular during complicated reasoning. This raises concerns on whether its decision can be trusted~\cite{smith2024llms}. Second, travel planning is a \emph{time-dependent} task that requires constantly re-planning due to ever-changing situations. Even with exactly the same request, the optimal itinerary may be different given varying prices and availability. Third, such a decision is highly \emph{personalized} depending on the private constraints and preferences. Users may speak a few brief words and expect the agent to give a solution that satisfies all of their implicit constraints, which can be quite subtle to capture from past conversations. 

\begin{figure*}
    \centering
    \includegraphics[width=0.9\textwidth]{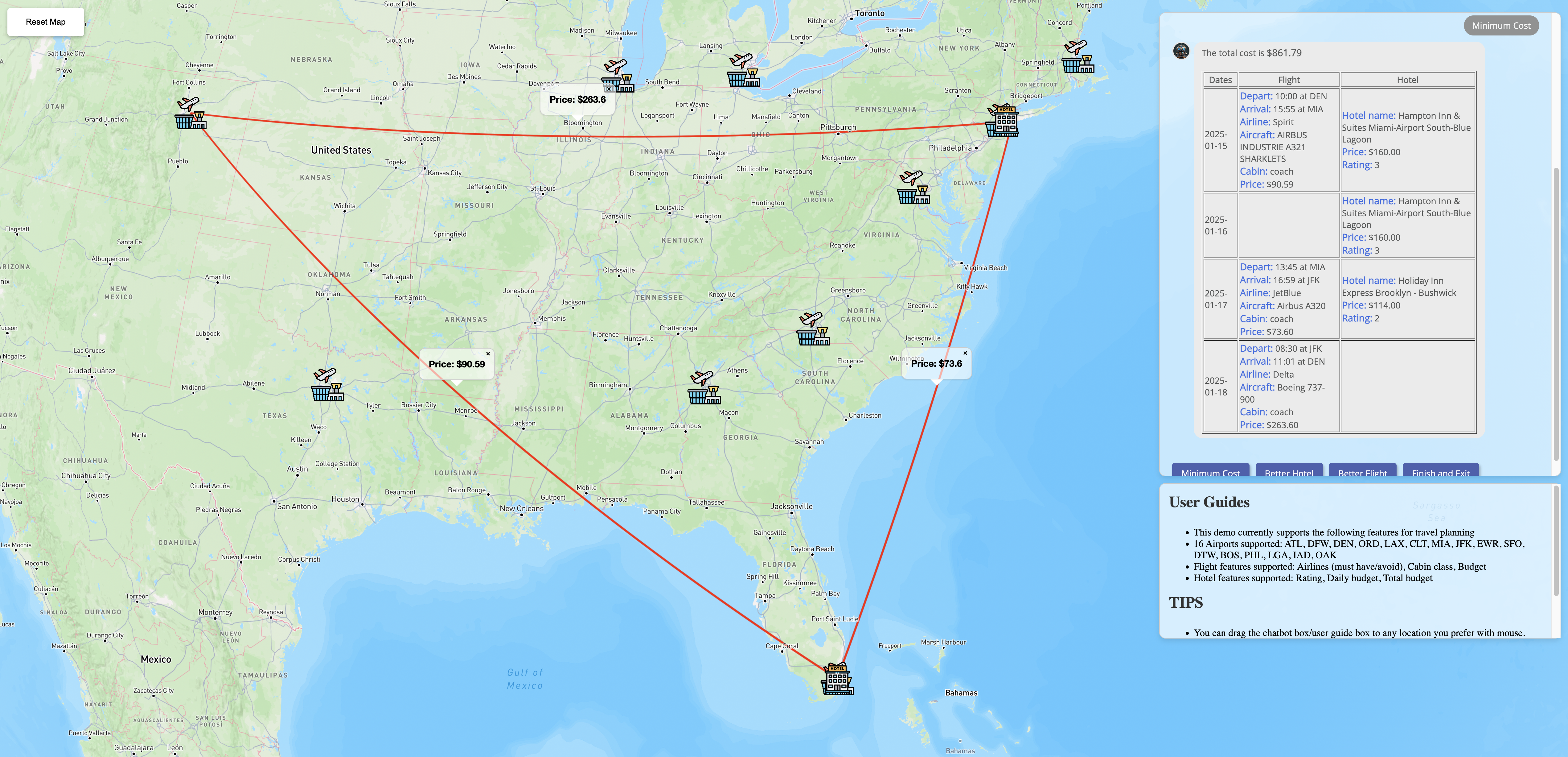}
    \caption{\small Front-end interface of \ours{}. Users send their natural language requests to the demo system (\ours{}), and \ours{} replies with itineraries that satisfy user constraints and is optimal with respect to various criteria (e.g., minimal cost).} 
    \label{fig:interface}
\end{figure*}

In this paper, we propose \ours{}, a demo system that takes natural language instructions from users and outputs optimal travel itineraries in seconds. To achieve this, our system leverages the power of LLMs and existing symbolic solvers, e.g., Mixed Integer Linear Programming (MILP). It first converts natural language instructions into a symbolic representation, which is solved by the symbolic solver, and then replies to the user with natural language outputs and a rich visualization. Compared to pure LLM-based systems (e.g., ClaudeAtlas\footnote{\url{https://devpost.com/software/kickass-team}}, Expedia Romie\footnote{\url{https://partner.expediagroup.com/en-us/innovation/labs}}), \ours{} provides an up-to-date, precise and executable itinerary with guarantees, in almost real time ($\sim 5$ seconds per request). 

\section{Related Works}
\textbf{LLM for reasoning/planning}. Teaching LLMs to do well in reasoning and planning tasks remains a challenging problem, even for SoTA LLMs~\cite{zheng2024natural,reid2024gemini}. Previous works like CoT~\cite{wei2022chain}, ToT~\cite{yao2024tree}, ReAct~\cite{yao2022react}, Reflexion~\cite{shinn2024reflexion}, using synthetic data~\cite{setlur2024rl,russell2016artificial}, and multi-agent frameworks~\cite{wu2023autogen,hong2023metagpt} improve the reasoning power of LLMs in complicated problems but still cannot guarantee feasibility and optimality~\cite{xie2024travelplanner,xie2024human}. More importantly, due to the blackbox nature of LLMs, it remains an open problem on understanding failure modes in reasoning~\cite{williams2024easy,chen2024premise}, let alone generate guaranteed results that can be trusted by users.

\noindent
\textbf{Hybrid System of LLM and Solvers}. Combining symbolic solvers with LLMs has been explored in many abstract planning (e.g,~\cite{liu2023llm+,silver2024generalized,silver2022pddl}) and real-world planning scenarios~\cite{tang2024synergizing}. 

For travel planning, ~\cite{hao2024large,de2024trip} show that prompt engineering in pre-trained language models can be used to generate code (or symbolic specification) to invoke symbolic solvers such as formal verification tools like SMT~\cite{bjorner2015nuz} solvers, or $A^*$~\cite{russell2016artificial}, to solve travel planning problems (e.g.,~\cite{xie2024travelplanner}). In contrast, our \ours{} chooses to focus on real-world travel planning that may last for multiple days with realistic constraints (Table~\ref{tab:travel-generator-specification}). \ours{} uses JSON format as symbolic specification because it has much simpler structures than generated codes, and can be guaranteed by constrained generation techniques using finite state machine (FSA)~\cite{geng2023grammar}, which makes self-consistency-based verification, benchmarking, and training easier (see Sec.~\ref{sec:experiments} for details). 
This also makes \ours{} independent of the specific solver (e.g., SCIP~\cite{bestuzheva2021scip}, Gurobi, etc.) and language used to solve the underlying MILP problem. Instead of prompt engineering, \ours{} also does model fine-tuning with thorough performance evaluation, including self-consistency and a thorough human study with $\sim 1.3$k participants, which is not provided in previous works.

\begin{table}
    \centering
    \small
    \adjustbox{max width=0.48\textwidth}{
    \begin{tabular}{ll}
       \toprule
       \textbf{Item}  &  \textbf{Description} \\
       \midrule
        \multirow{4}{*}{Airline Constraints} & price range, (soft) departure \\
        & and arrival constraints, cabin class, \\
        & refundablity, non-stopness, \\
        & plane type, airline preferences. \\
       \midrule
        Hotel Constraints & price range, rating, brands. \\
       \midrule
        Budget Constraints & Total budget, everyday budget.\\
        \bottomrule 
    \end{tabular}
    }
    \caption{\small Factors considered in travel request generation.} 
    \label{tab:travel-generator-specification}
\end{table}

\begin{figure*}[t]
    \centering
    \includegraphics[width=\textwidth]{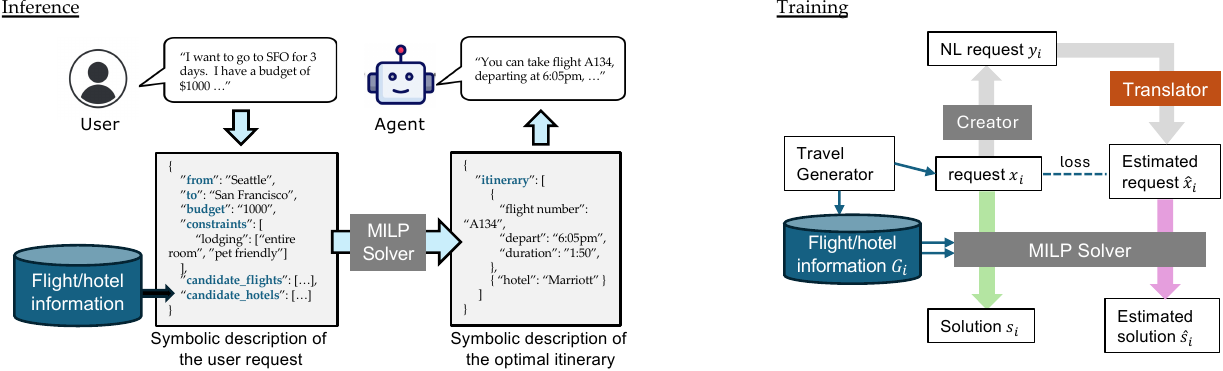}
    \caption{\small Overview of the workflow of \ours{}. \textbf{Inference:} our system first translates the user travel request in natural language (NL) into the symbolic description of a Mixed Integer Linear programming (MILP) solver using a fine-tuned Large Language Model (LLM), calls the solver to find its optimal solution that satisfies all constraints, and then returns the itinerary in natural language. \textbf{Training}: \ours{} has three components. A \emph{Travel Generator} that generates flight/hotel information training data based on real-world data, and symbolic user request $x_i$. An \emph{Instruction Translator} a pre-trained LLM fine-tuned to translate the NL user request $y_i$ to its symbolic form $\hat x_i$, learned by self-consistency between the groundtruth request $x_i$ and the estimated user request $\hat x_i$. A \emph{Travel solver} that solves the estimated symbolic request $\hat x_i$ and yields the estimated solution $\hat s_i$.}
    \label{fig:overview}
\end{figure*}

\section{Overview of \ours{}}
Fig.~\ref{fig:interface} shows the front-end interface of \ours{}. Users can obtain travel itineraries in a few seconds by sending natural language requests in the semi-transparent dialog box. Users can also visualize candidate itineraries on the rendered map of the globe, and select based on their preferences. We use a hybrid design  leveraging both LLMs and symbolic solvers that can deal with natural language input and still guarantee the feasibility and optimality of the output itineraries, if user requests are translated correctly by the fine-tuned LLM.  

Fig.~\ref{fig:overview} shows the overview of the \ours{} workflow. The components are: (1) A \emph{Symbolic Travel Generator} which generates available flight and hotel information $G_i$ using existing real-world data as well as symbolic user requests $x_i$ (both in JSON format) where $i$ is the sample index. (2) \emph{Instruction Creator and Translator} that converts a user request $x_i$ from JSON to a natural language (NL) request $y_i$, and a translator to convert the NL requests $y_i$ back to its symbolic form in JSON $\hat x_i$ (Sec.~\ref{sec:travel-gen}). (3) \emph{Travel Solver}, a Mixed Integer Linear Programming (MILP) solver that solves the underlying combinatorial optimization in its symbolic form, parameterized by $(x_i, G_i)$, and gives the optimal solution $s_i$. That is, $s_i = \arg\min_{s'} f(s'; x_i, G_i)$, where $f > 0$ is the cost function to be minimized. During the user interaction, the solver only has access to an estimate of the user request $\hat x_i$, and the corresponding solution $\hat s_i = \arg\min_{s'} f(s'; \hat x_i, G_i)$.

\iffalse
\item \emph{\st{Solution Replier} \maryam{Output Generator?}}. The optimal solution $s_i$ (or $\hat s_i$) is then converted back to a natural language response $r_i$ (or $\hat r_i$). We also train a Reply Translator $\phi$ that converts the natural language back to the solution. \maryam{Why do we need a Reply Translator?}
\fi

\section{Methodology}
\subsection{Symbolic Travel Generator}
\label{sec:travel-gen}
Since the existing TravelPlanner dataset~\cite{xie2024travelplanner} has a limited number of samples and does not provide symbolic grounding of user requests, we created our own Travel Generator that generates user requests and the corresponding flight and hotel information in symbolic form.  

\textbf{Travel Request}. We consider a variety of variables when generating travel requests (see Table~\ref{tab:travel-generator-specification} for a complete list). We mostly consider round trips of 2 or 3 cities ($1$ or $2$ stops) over multiple days (include <5\% one-way for diversity). We randomly sample values for the constraints in Table~\ref{tab:travel-generator-specification} and prompt Llama-3 70B~\cite{llama3modelcard} to convert the symbolic representation into natural language. We generate $238$k training samples and $29.8$k test samples.

As is common in synthetic data generation with LLMs~\cite{Ji2022SurveyOH}, there was some degree of inconsistency between the symbolic representation and generations, primarily in the ordering of departure and return dates. We again prompted Llama-3 70B to filter samples with this inconsistency as a few-shot task, removing approximately $27\%$ of samples leaving $173.7$k training and $21.8$k test samples.

\textbf{Generated Flight and Hotel information $G_i$}. We use the flight price dataset\footnote{\url{https://www.kaggle.com/datasets/dilwong/flightprices}} which contains existing real-world, one-way US domestic flight information from Expedia from Apr. 16, 2022 to Oct. 5, 2022 to build our travel generator. We replicate the data to cover a longer time frame. For hotels, we include public information and then add noise to prices, departure/arrival dates, and other attributes. We combine the two to create synthetic flight and hotel information $G_i$. 

\def\ev{\mathrm{ev}}
\def\src{\mathrm{src}}
\def\dst{\mathrm{dst}}
\def\dep{\mathrm{dep}}
\def\land{\mathrm{land}}
\def\air{\mathrm{air}}
\def\checkin{\mathrm{ckin}}
\def\checkout{\mathrm{ckout}}

\subsection{Travel Solver}
We build a combinatorial solver to compute optimal solutions to the MILP formulation of the travel planning problem using SCIP~\cite{bestuzheva2021scip}. We discretize the time into $T$ slots, over the travel span (e.g., 3 days). A traveller is at location $l$ at time $t$ if and only if the corresponding variable $u_{l}(t) = 1$. The traveller must maintain location continuity and cannot teleport unless some event happens: $e(t) = 0 \Rightarrow u_l(t+1) = u_l(t)$. A traveller may be sleep at time slot $t$, which is represented as $m(t) = 1$. A hotel $j$ (or a flight $j$) is booked if $h_j = 1$ (or $f_j = 1$). All the variables are binary.  

To make sure the resulting solution is feasible, we impose the following three types of constraints.   

\textbf{Commonsense constraints}. The traveller can only be present at a single location at time $t$, which means $\sum_l u_{l}(t) = 1$. The traveller needs a minimal $L$ time slots per day, which can be represented as $\sum_{t\in [\mathrm{day\ evening}]} m(t) \ge L$.  

\textbf{Flight constraints}. If the traveller takes the flight $j$ (i.e., $f_j = 1$) that departs from location $\mathrm{src}$ to location $\mathrm{dst}$, then the following constraints should be satisfied:
\begin{equation}
\small
    f_j = 1 \Rightarrow \left\{\begin{array}{c}
    u_\src(t_\dep) = 1, u_\air(t_\dep + 1) = 1 \\
    u_\dst(t_\land) = 1, u_\air(t_\land - 1) = 1 \\
    e(t_\dep) = e(t_\land - 1) = 1
    \end{array}\right. \label{eq:flight-constraint}
\end{equation}
where, $t_\dep$ and $t_\land$ are the departure and landing time slots, and $e(t)$ is a binary variable suggesting whether there is an event happening at time slot $t$.  

\textbf{Hotel constraints}. If the traveller decided to reside in hotel $j$ (i.e., $h_j=1$) at location $l$, then the following constraints need to be satisfied:
\begin{equation}
\small
    h_j = 1 \Rightarrow \left\{\begin{array}{c}
    u_l(t_\checkin:t_\checkout) \ge m(t_\checkin:t_\checkout) \\
    m(t_\checkin:t_\checkout)\mathrm{\ allowed\ to\ be\ 1} 
    \end{array}\right. \label{eq:hotel-constraint}
\end{equation}
where, $t_\checkin$ and $t_\checkout$ are the earliest and latest check-in and check-out times for hotel $j$. 

\textbf{Encoding (``implies'' $\Rightarrow$) conditions}. Note that MILP is able to encode conditional constraints (e.g., Eqn.~\ref{eq:flight-constraint} and Eqn.~\ref{eq:hotel-constraint}). Please check Appendix~\ref{sec:conditional-constraints} for details.  

\section{Experiments}
\label{sec:experiments}
\subsection{Automatic Evaluation by Self-consistency}
\textbf{Quality of Instruction Translator}. We evaluate the quality of the generated symbolic form $\hat x_i$ from the Translator, by comparing with the original symbolic form $x_i$ that is used to generate the natural language request $y_i$. 

%\textbf{Evaluation metric}.
To compare the original symbolic user request $x_i$ and reconstructed request $\hat x_i$ (both in JSON) from natural language request $y_i$, we use \emph{exact match} (EM) accuracy that scores $0$ if any of the entries in the two JSONs do not match. Additionally, since the Translator is generating output structured as JSON, we use vLLM logits\_processors to ensure the model output is properly structured~\cite{kwon2023efficient}. We refer to this as Constrained Decoding.

In Table~\ref{table:test-loss}, we report exact match accuracy and validity of the output as JSON for both Constrained and Unconstrained Decoding on the test set. With constrained decoding, the Translator achieves $92.0\%$ exact match accuracy with output being valid JSON $100\%$ of the time (because we forced it to be). Unconstrained decoding is surprisingly close to constrained decoding with an EM of $91.2\%$ and produces valid JSON $99\%$ of the time. We find that the filtering step discussed in Sec.~\ref{sec:travel-gen} to be critical for unconstrained decoding to produce valid JSON at such a high degree. Constrained decoding is roughly $10\%$ slower than unconstrained decoding but the $1\%$ failure rate leads to a worse user experience,s so we deploy constrained decoding in the demo.

\begin{table}
\centering
\small
  \begin{tabular}{ccc}
    \toprule
      \textbf{Decoding} & \textbf{EM Accuracy} & \textbf{Valid JSON} \\ \midrule
      Constrained &  $92.0\%$ & $100.0\%$ \\ 
      Unconstrained & $91.2\%$ & $99.1\%$\\
    \bottomrule 
  \end{tabular}
\caption{\small Exact Match accuracy and validity of generations as JSON of \ours{} with Constrained and Unconstrained decoding on $21.8$k test samples.}
\label{table:test-loss}
\end{table}

Table~\ref{table:by-constraints} provides a breakdown of the errors and number of samples by the number of hotel constraints, airline constraints and cities. We point out that EM accuracy \textit{decreases} as the number of airline constraints \textit{increases} but is relatively robust across the number of hotel constraints and cities. We hypothesize the decreasing performance with airline constraints is due to data imbalance (i.e., there are only $173$ samples with 8 constraints versus $9777$ with 5 constraints) which can be addressed by changing the sampling parameters during data generation.  

\begin{table}[ht]
  \centering
  \small
  \adjustbox{max width=0.48\textwidth}{
    \begin{tabular}{cccccc}
        \toprule
        \textbf{Hotel Constraints} & \textbf{2} & \textbf{3} & \textbf{4}\\
        \midrule
        \textbf{EM Accuracy} &$91.5\%$& $92.5\%$& $91.7\%$\\ 
        \textbf{\# samples} & $3345$ & $10438$ & $8001$ \\
        \midrule
        \textbf{Airline Constraints} & \textbf{4} & \textbf{5} & \textbf{6} & \textbf{7} & \textbf{8}\\
        \midrule
        \textbf{EM Accuracy} &$95.9\%$& $92.8\%$& $91.1\%$ & $77.0\%$& $78.6\%$\\ 
        \textbf{\# samples} & $4974$ & $9777$ & $5555$ & $1299$ & $173$\\
        \midrule
        \textbf{Cities} & \textbf{2} & \textbf{3} \\
        \midrule
        \textbf{EM Accuracy} &$91.9\%$& $93.0\%$& \\ 
        \textbf{\# samples} & $18998$ & $2786$ \\
        \bottomrule
    \end{tabular}
  }
\caption{\small Exact Match (EM) accuracy of \ours{} and the number of samples when sorting by the number of hotel constraints, airport constraints and cities. Accuracy decreases as the number of airline constraints increases but is relatively robust across the number of hotel constraints and cities.}
\label{table:by-constraints} 
\end{table}

\begin{figure}
    \includegraphics[width=0.48\textwidth]{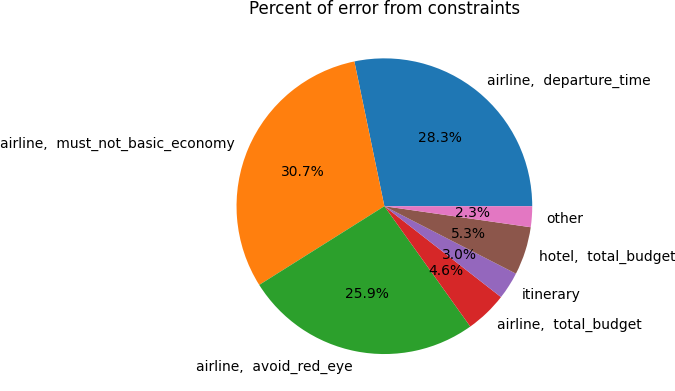}
    \caption{\small The breakdown of sources of error in EM accuracy. The three major sources of error are the airline constraints \texttt{must\_not\_basic\_economy}, \texttt{departure\_time}, and \texttt{avoid\_red\_eye}.} 
    \label{fig:pie_error}
\end{figure}

Fig.~\ref{fig:pie_error} provides a breakdown of the sources of error of our model. The three major sources are the airline constraints \texttt{must\_not\_basic\_economy}, \texttt{departure\_time}, and \texttt{avoid\_red\_eye}. A manual inspection reveals that Llama-3 is somewhat insensitive to these constraints and a common failure mode is that they simply do not appear in the generated NL requests. To further filter for these samples, as we did with issues with departure and return dates discussed above, is left for future work.

\noindent
\textbf{Quality of solutions}. When there is no exact match, we instead evaluate the end-to-end performance by checking the feasibility and optimality of the response $\hat s_i$, by checking the \emph{quality ratio} of the \textit{ \it cost} $f(\hat s_i; x_i, G_i)$ of generated solution $\hat s_i$ (as a function of estimated user request $\hat x_i$), to the minimal cost $f(s_i; x_i, G_i)$ if the solver is fed with a groundtruth user request $x_i$. Note, $\hat s_i$ is computed by solving the \textit{estimated} symbolic user request $\hat x_i$ but we evaluate the cost with respect to the ground truth $x_i$.
\begin{equation}
\small
    score(i) = f(s_i; x_i, G_i)/f(\hat s_i; x_i, G_i)
\end{equation}
Since $f(\hat s_i; x_i, G_i) \ge f(s_i; x_i, G_i)$, we know $0 \le score(i) \le 1$ where a score of $0$ corresponds violating one or more constraints and $1$ corresponds to the optimal solution. A score between $0$ and $1$ corresponds to finding sub-optimal solutions to some constraints. We partition the $21.8$k test samples into $8$ subsets of $2.7$k samples. The mean and standard deviation for \ours{} over the $8$ subsets is $0.979\pm0.002$, which is very close to $1$ (optimal). Within the samples where the generated constraints are not an exact match, the score is $0.726 \pm .0234$. 

\begin{table}[h]
  \centering
  \small
  \adjustbox{max width=0.52\textwidth}{
    \begin{tabular}{lc}
        \toprule
        \textbf{Response phase} & \textbf{Time (s)}\\
        \midrule
        Instruction Translator & 2.508\scriptsize{$\pm$0.116}\\
        MILP Solver &  \\
         \ - Loading constraints  & 0.047\scriptsize{$\pm$0.061}\\
         \ - Solving  & 0.527\scriptsize{$\pm$0.457}\\
         \ - Total & 0.575\scriptsize{$\pm$0.507}\\
        \bottomrule
    \end{tabular}
  }
  \caption{\small Time spent on each phase of \ours{}. We report the average and standard deviation over 100 examples.} 
\label{table:speed}
\end{table}

% \vspace{5pt}

% \vspace{-0.05in}
\subsection{Efficiency of \ours{}} 
% \vspace{-0.05in}
We also evaluate the performance of \ours{} by profiling the two major components: generation speed of the Translator and the speed of the MILP solver, tested on a AWS P4de node using one A100 for LLM inference and one CPU (Intel Xeon Platinum 8275CL@3GHz) core for the solver. As shown in Table~\ref{table:speed}, the primary bottleneck in our system is the model inference cost which takes $81.3\%$ of the compute time. Overall, \ours{} is light-weight and provides responses in real-time.

\subsection{Human evaluation} 

We performed an online survey and qualitative interviews to collect human judgment and feedback about our system's performance. The goal of the human evaluation study was two-fold: (1) validate performance and subjective perception of our system's outputs through a large pool of lay-people who routinely travel, and (2) identify factors that contribute to perceived itinerary quality to inform future work.

We screened from a broad pool of US-based participants who travel four or more times per year to complete a survey evaluating model performance. To maximize evaluation coverage, we randomly sampled $50$ natural language travel queries from our generated test set, stratifying by number of stops ($60\%$ one-/$40\%$ two-stop) and encompassing a variety of trip durations and budgets. We then ran the queries through \ours{}, rendering the map and detailed travel itinerary per trip presented in tabular form (see Fig.~\ref{fig:interface}) via a chat interface. We randomly assigned each of the $1385$ participants to $5$ of the sampled query-itinerary pairs and ask them to evaluate along three axes (see below). In addition, we also ask the participants to rank the factors that affect their travel decisions, and conduct in-depth interviews to find ways to improve \ours{} (see Appendix \ref{ref:user-study} for details).

\subsubsection{On Satisfaction, Value and Efficiency}
For each query-itinerary pair, participants answered three questions: (1) how much the query was satisfied, (2) the value and (3) the efficiency of the itinerary. Participants noted that they require extensive comparison on many hard (e.g., price) and soft (e.g., aesthetic) criteria as part of assessing optimality, often over many hours of research. Consequently, measuring whether a given itinerary was optimal via human evaluation was determined infeasible. Therefore, we use subjective metrics like (2) and (3) as proxy evaluations for the optimality of each itinerary, absent being able to assess optimality via human evaluation.  

We evaluated the survey responses by computing a score constructed similarly to Net Prompter Scores (NPS~\cite{fisher2019good}). This system used a five-point scale (percentage of supporters minus detractors where 5s are coded as promoters and 1-3s as detractors), as shown in Table ~\ref{table:nps}. Our primary \emph{`satisfies the request'} question received a {40.0}. Our secondary \emph{`value'} and \emph{`efficiency'} questions scored {35.1} and {36.9}, respectively. Overall, we consider these promising results, indicating user acceptance on all three evaluation metrics. We note that while this evaluation does not use the original NPS language, the method of analysis still enables us to understand the relative proportion of respondents who view our model favorably. Additionally, no material difference is seen between user evaluations of the one- and two-stop itinerary.

\begin{table}[h]
  \centering
  \small
  \adjustbox{max width=0.48\textwidth}{
    \begin{tabular}{lc c |c }
        \toprule
        \textbf{Question} & \textbf{Detractors \%}\ & \textbf{Promoters \%}\ & \textbf{Net \%}\\
        \midrule
        ...fully satisfies the...request & -13.3 & +53.3 & +40.0\\
        ...offers good value for the money... & -16.8 & +52.0 & +35.1\\
        ...is efficient... & -16.2 & +53.1 & +36.9\\
        \bottomrule
    \end{tabular}
  }
  \caption{\small Net Prompter-like Score (NPS) and its breakdown in survey questions. Please check the complete form of the questions (as well as other details) in Appendix \ref{ref:user-study}.} 
\label{table:nps}
\end{table}

\subsubsection{Preference ranking}
Price and preferred travel times were ranked as the most important criteria in trip assessment, reinforcing the selection of these proxy criteria. We see these preferences manifesting in at least two large and distinct user clusters: the first group includes price sensitive travelers, looking for high value; the second cares more about departure times, service levels, and brands. Future work may include personalization; we expect closer alignment to user optimality by inferring user groups to re-weight criteria before computing itineraries. 

\subsubsection{In-depth Interviews}
We then conducted in-depth user interviews with {8} participants who matched the recruitment criteria for the survey. These interviews followed a semi-structured, in-depth format. Participants were asked to reflect on recent travel, walking through their tools used, process of searching for and selecting flights and accommodations, including points of high and low friction and heuristics for prioritization. Finally participants assessed stimuli, which were generated via the same criteria as used to populate the survey.

Together, the survey and user interviews illuminated the following themes for future improvement. \textbf{Prioritization}. User requests demonstrate a hierarchy of importance, e.g., flight selection often precedes hotel bookings. \textbf{Flexibility}. Trip details should be changed with ease and enable comparison. \textbf{Personalization}. Users have a variety of preferences, e.g., cheap vs. cozy, business vs. casual, family vs. solo trips, etc. Many of them are implicit. Moreover, special needs like ``\emph{The room door opens to a hallway}'' may not be available but can be part of user's ideal selection criteria. \textbf{Trust of AI agents}. Decisions made by the agent should be readily verifiable by users as feasible, optimal and fit to their personal use cases. For this, more convenient tools are needed to visualize copious information for confidence boost. While \ours{} moves towards these goals (e.g., guaranteed quality of solutions by solver), more works can to be done.

\section{Conclusion and Future Work}
In this work, we propose \ours{}, and end to end system which plans travel itineraries from user requests in natural language. \ours{} uses a hybrid architecture that combines an LLM with combinatorial solvers, dynamically formulating travel requests into a well-defined MILP problem, and translating the solution computed by the solver back to natural language. Overall, the system responds almost in real-time ($\sim 5$ seconds), and outputs feasible and optimal guaranteed travel itineraries, given correctly understood user requests by the fine-tuned LLM, which happens $>90\%$ of the time for queries up to $6$ airline constraints and up to $4$ hotel constraints. For this, a data generation pipeline is developed to provide synthetic symbolic and natural language pairs for model training. 

We recognize that achieving true optimality requires a system that enables robust personalization, and human-driven filtering and selection. As a result, we anticipate the need for a human benchmark task that enables respondents to stipulate a travel goal in real time and compare between a few near-optimal results, both to measure system performance and to collect signal for improvement. Future developments will therefore explore multi-round dialog and personalization to further improve user experience, and end-to-end trainable pipelines to make the system more adaptive. 

\bibliography{custom}

\clearpage
\onecolumn

\appendix
\section{Using MILP solver to encode conditional constraints}
\label{sec:conditional-constraints}
Suppose $\{z_j\}$ are binary variables, then the conditional constraint ``if all $z_j = 1$, then $x = y$'' can be formulated as the the following:
\begin{equation}
    x \le y + M \sum_j (1 - z_j), \quad\quad y \le x + M \sum_j (1 - z_j)
\end{equation}
where $M$ is a big constant. Intuitively, if all $z_j = 1$, then the above two constraints are equivalent to $x \le y$ and $y \le x$, which is $x = y$; if $k$ of the binary variable $\{z_j\}$ are zero, then the above two constraints become $x \le y + k M$ and $y \le x + k M$, which becomes trivial for big $M$. 

\section{Details of the User Study}
\label{ref:user-study}

1. Survey Design

Q1.  [RANK]- What matters most to you when selecting a travel itinerary (airfare and hotels)?
•	Total Price
•	Value per dollar
•	Minimal Time in Transit
•	Simple or Few Steps
•	Travel/stay with preferred brands
•	Travel at preferred times
•	Travel at specific level of service (e.g. hotel stars, airfare class)

Q2-Q6.  [SCALE]- For the following question, please reference the image shown. How much do you agree or disagree with the following statements? (5 Point Scale: Strongly Disagree - Strongly Agree) (Repeated 5 times)

•	This travel itinerary fully satisfies the corresponding travel request.
•	This travel itinerary is efficient, given the corresponding travel request.
•	This travel itinerary offers good value for the money, given the corresponding travel request.

Q7.  [OPEN END]- How could the format or quality of these itineraries be improved?

\section{Details of \ours{} Demo}\label{sec:features}

We introduce the key features of our demo in detail, using the same example as shown in Fig.~\ref{fig:interface}.

\textbf{User request.} The user request in our example is ``\textit{Embark on a thrilling journey with these requirements. Flights: coach class, non-stop, no basic economy or mixed cabin, with a total budget of \$1383. Hotels: daily budget \$317, total budget \$952. Travel dates: January 15th, 2025, DEN to MIA, January 17th, 2025, MIA to JFK, and January 18th, 2025, JFK to DEN. The adventure awaits!}''

\begin{figure*}[h]
    \centering
    \includegraphics[width=0.6\textwidth]{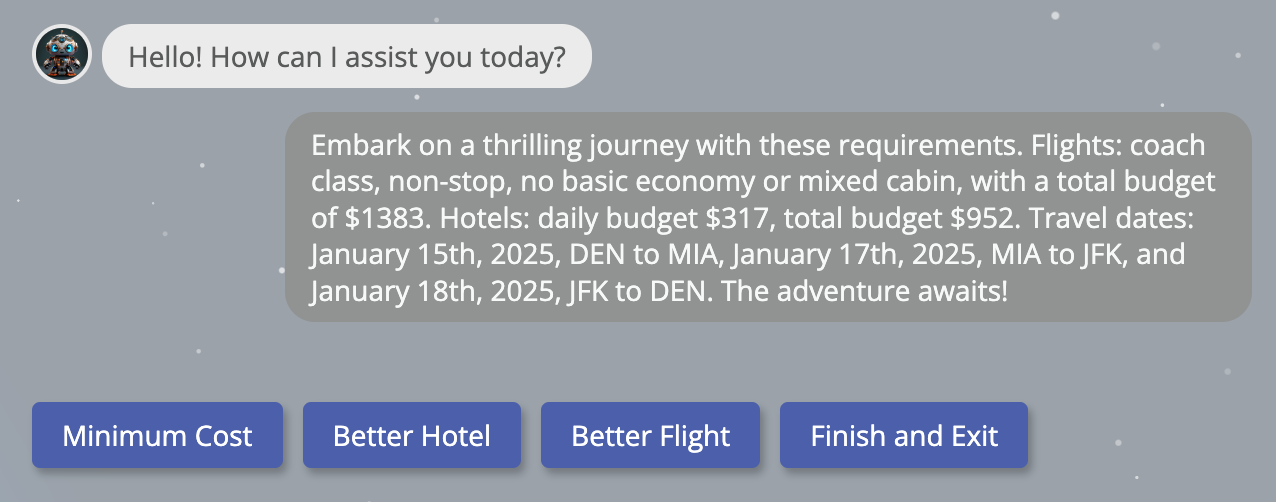}
    \caption{Itinerary options in \ours{} demo.} 
    \label{fig:options}
\end{figure*}

\textbf{Itinerary Options.} For a user travel request, \ours{} gives three itinerary options with three different considerations: 1) \emph{Minimum Cost:} the total cost (flights+hotels) is minimized; 2) \emph{Better Hotel:} More tolerant of hotel costs for a better hotel experience; and 3) \emph{Better Flight:} More tolerant of flight costs for a better flight experience. These options are materialized by different objectives in the MILP travel solver. We show the user interface of three itinerary options in Fig.~\ref{fig:options}.

\begin{figure*}[h]
    \centering
    \includegraphics[width=\textwidth]{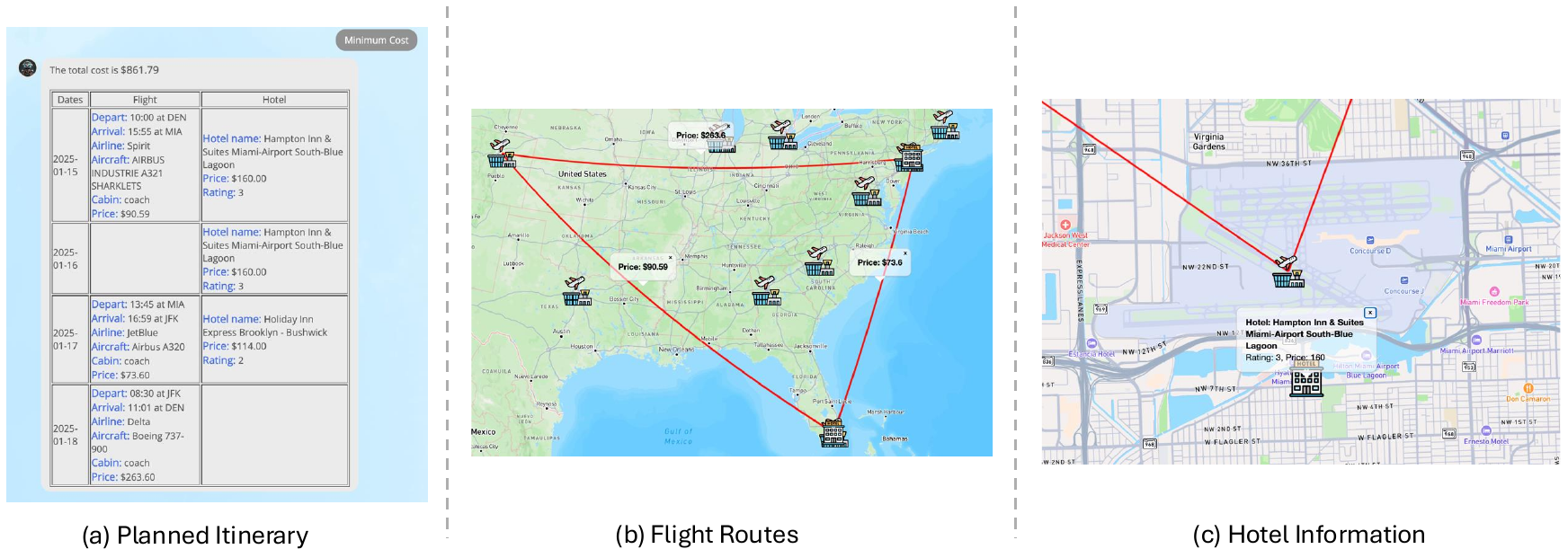}
    \caption{Details of demo. (a) Planned itinerary is shown in tabular view; (b) Flights routes are shown on the map with prices on each travel segment; (c) Hotel infomation, including name, rating and price.} 
    \label{fig:demo_screenshots}
\end{figure*}
\textbf{Planned Itinerary.} Fig.~\ref{fig:demo_screenshots} (a) showcases the planned itinerary with minimum cost as objective. \ours{} presents this itinerary in a tabular format, detailing the total budget, flight specifics, and hotel information.

\textbf{Flight Routes.} As shown in the detailed view in Fig.~\ref{fig:demo_screenshots} (b), \ours{} presents a sequence of flights according to the user's request (DEN to MIA, MIA to JFK, and JFK to DEK), with the corresponding prices of flights hovering above each each route.

\textbf{Hotel Information.} Once clicking the hotel icon, \ours{} provides a zoomed-in view of the suggested hotels with their ratings and prices. For instance, as shown in Fig.~\ref{fig:demo_screenshots} (c), \ours{} has booked the "Hampton Inn \& Suites Miami-Airport South-Blue Lagoon" for the user's stay in Miami (MIA). This selection meets the user's daily hotel budget constraint of \$317. Note that if a user specifies a minimum hotel rating, the MILP solver in \ours{} ensures this requirement is also met.

\textbf{Packages Acknowledgement.} Our \ours{} demo is built upon Mapbox~\footnote{\url{https://www.mapbox.com/}} and BotUI~\footnote{\url{https://github.com/botui/botui}}.

\end{document}